\documentclass[preprint,12pt]{elsarticle}

\usepackage{amssymb}
\usepackage{amsmath}
\usepackage{amsmath,amssymb,amsfonts}
\usepackage{algorithmic}
\usepackage{graphicx}
\usepackage{textcomp}
\usepackage{xcolor}
\usepackage{multirow}
\usepackage{subcaption}
\usepackage{booktabs}      
\usepackage{colortbl}      
\usepackage{hyperref}
\usepackage{cleveref}
\hypersetup{
hidelinks,
colorlinks=true,
linkcolor=red,
citecolor=blue
}

\definecolor{mypink}{RGB}{255,182,193} 

\journal{Expert Systems with Applications}

\begin{document}

\begin{frontmatter}



\title{Decoding Visual Neural Representations by Multimodal with Dynamic Balancing}


\author[Durham]{Kaili~Sun}\ead{kaili.sun@durham.ac.uk}
\author[Durham]{Xingyu~Miao$^{\#}$}\ead{xingyu.miao@durham.ac.uk}
\author[Northumbria]{Bing~Zhai}\ead{bing.zhai@northumbria.ac.uk}
\author[THU]{Haoran~Duan}\ead{haoranduan28@gmail.com}
\author[Durham]{Yang Long\corref{mycorrespondingauthor}}\ead{yang.long@durham.ac.uk}
\address[Durham]{Department of Computer Science, Durham University, UK.}
\address[Northumbria]{Computer and Information Sciences, Northumbria
University, UK.}
\address[THU]{Department of Automation, Tsinghua University, China.}
\cortext[mycorrespondingauthor]{Corresponding author $^{\#}$ equal contribution}


\begin{abstract}
In this work, we propose an innovative framework that integrates EEG, image, and text data, aiming to decode visual neural representations from low signal-to-noise ratio EEG signals. Specifically, we introduce text modality to enhance the semantic correspondence between EEG signals and visual content. With the explicit semantic labels provided by text, image and EEG features of the same category can be more closely aligned with the corresponding text representations in a shared multimodal space. To fully utilize pre-trained visual and textual representations, we propose an adapter module that alleviates the instability of high-dimensional representation while facilitating the alignment and fusion of cross-modal features. Additionally, to alleviate the imbalance in multimodal feature contributions introduced by the textual representations, we propose a Modal Consistency Dynamic Balance (MCDB) strategy that dynamically adjusts the contribution weights of each modality. We further propose a stochastic perturbation regularization (SPR) term to enhance the generalization ability of semantic perturbation-based models by introducing dynamic Gaussian noise in the modality optimization process. The evaluation results on the ThingsEEG dataset show that our method surpasses previous state-of-the-art methods in both Top-1 and Top-5 accuracy metrics, improving by 2.0\% and 4.7\% respectively.
\end{abstract}


\begin{highlights}
\item A novel HMAVD fuses EEG, image, and text to boost visual decoding.
\item Text bridges EEG and image, stabilizing high-dim features for alignment.
\item Dynamic MCDB and SPR balance modality gradients, boosting robustness.

\end{highlights}


\end{frontmatter}



\section{Introduction}
Decoding visual neural representations is a key approach to understanding the mechanisms underlying human perception and cognition \cite{luu2016gait, ke2020online, ding2021repvgg,miao2025rethinking,miao2024dreamer}. Among the various neuroimaging techniques, EEG has emerged as an important tool for decoding brain activity evoked by visual stimuli, owing to its high temporal resolution, portability, and non-invasive nature \cite{liao2014decoding, tariq2018eeg, craik2019deep}. However, the high noise levels and low spatial resolution of EEG signals present significant challenges for precise decoding \cite{mullen2015real}.

Recent advances in deep learning have provided new solutions for processing EEG signals \cite{zhang2019making, schirrmeister2017deep}, nevertheless, these methods still exhibit limited generalization capabilities when addressing zero-shot tasks involving unseen categories. An effective strategy to overcome the challenges posed by zero-shot tasks is to develop decoding models that generalize to unseen categories, which typically necessitates the integration of multimodal information to enhance cross-modal decoding performance. For instance, EEG Conformer \cite{song2022eeg} employs convolutional modules to extract local features from EEG signals and leverages Transformer models to capture global dependencies, thereby facilitating efficient alignment and fusion between visual stimuli and EEG data. In order to reduce reliance on large-scale annotated datasets, NICE \cite{song2023decoding} introduces a self-supervised learning strategy to further enhance performance in cross-modal decoding tasks. However, these approaches focus solely on mapping individual visual stimuli to EEG signals.

\begin{figure}[ht]
    \centering
    \includegraphics[width=0.7\textwidth]{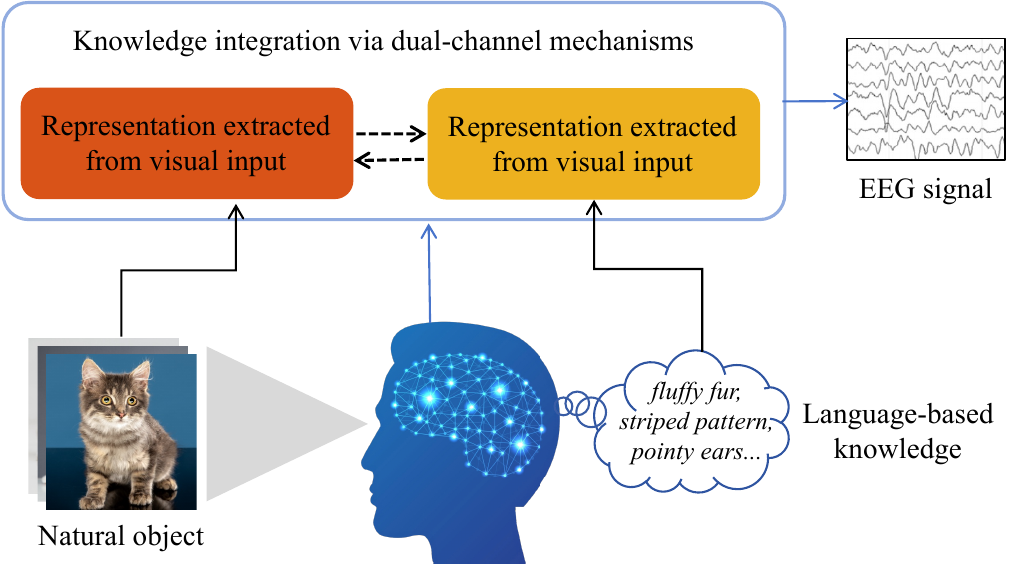}
    \caption{Dual-channel integration of knowledge in the human brain: When the human eye sees an object, the brain simultaneously processes visual input to extract representations and retrieves language-based knowledge as prior semantic context. These two channels interact to shape the brain's integrated understanding, as reflected in EEG signals, with linguistic information contributing to the refinement of vision-derived representations.}
    \label{fig:figure1}
\end{figure}

EEG features are often influenced by a combination of multiple modalities. Li et al. \cite{li2022eeg} integrated visual, auditory, and corresponding tactile stimuli using a stacked ensemble learning approach to improve the accuracy of emotion recognition from EEG data. Similarly, Duan et al. \cite{duan2023dewave} employed visual stimuli to induce internal speech, thereby augmenting the speech-related features within EEG signals. This joint visual-auditory stimulation notably enhanced the accuracy of EEG-based speech recognition. According to dual-coding theory \cite{paivio2013imagery, bi2021dual}, the formation of visual representations in the human brain is closely linked to linguistic prior knowledge (as shown in Figure \ref{fig:figure1}). BraVL \cite{du2023decoding} leverages this principle by fusing textual semantics with visual features in a joint learning framework, thereby improving the discrimination of visually similar categories. Compared to models relying solely on the visual modality, BraVL enhances the accuracy of decoding new visual categories from EEG.

These approaches demonstrate that multimodal fusion can effectively mitigate the limitations inherent to single modalities. However, they often assume balanced alignment between modalities and overlook potential misalignment issues arising from heterogeneity and distributional differences. Such misalignment is especially problematic under joint training strategies, where one modality may dominate the shared representation space, diminishing the contributions of other modalities and potentially leading to performance that is inferior to that of unimodal models \cite{wang2020makes, peng2022balanced}. Some studies \cite{wang2020makes, du2021improving, li2024uni, duan2023dynamic,duan2025parameter,miao2024ctnerf} have addressed this issue by designing modality-specific subnetworks via expert mixture architectures or by incorporating additional unimodal classifiers to support multimodal training. However, these methods inevitably incur extra computational overhead due to training additional neural modules.

\begin{figure*}[htbp]
    \centering   
    \includegraphics[width=\textwidth]{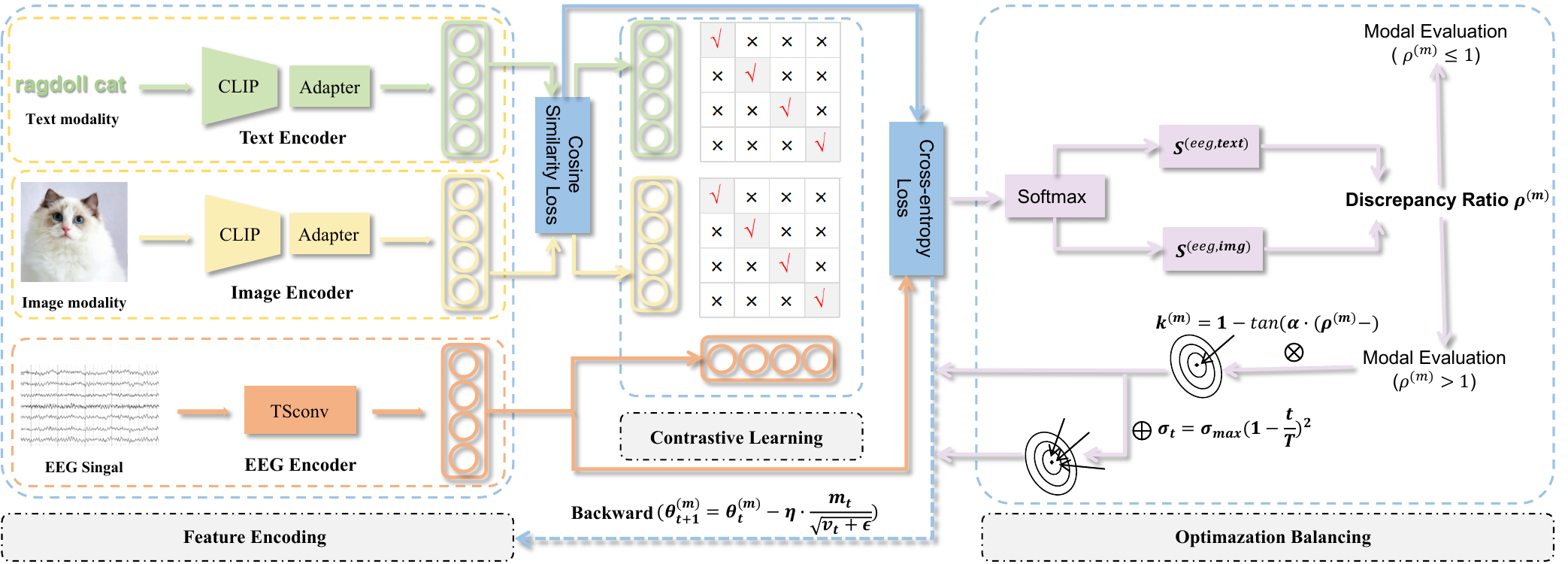}
    \caption{Overall Framework for multimodal EEG-Image-Text Alignment: During training, EEG signals, images, and text are processed by their respective encoders, with text optimized using CLIP and an adapter. Contrastive learning aligns multimodal embeddings by maximizing semantic consistency. The Optimization Balancing module, incorporating Modality Consistency Dynamic Balance and Stochastic Perturbation Regularization, ensures balanced contributions and enhances robustness. During testing, unseen images and text descriptions are pre-encoded as templates, and the model matches test EEG data to these templates for object recognition.}
    \label{fig:Multi-model Pre-training}
\end{figure*}

In this work, we propose a Harmonic multimodal Alignment for Visual Decoding, HMAVD (the overall framework is depicted in Fig. \ref{fig:Multi-model Pre-training}), aimed at improving the accuracy of decoding unseen visual categories from brain activity. To overcome the limitations in EEG decoding accuracy caused by insufficient visual modality information, we incorporate text as a semantic bridge to enhance cross-modal alignment and fusion. Moreover, to prevent the dominant modality from suppressing the contributions of others during joint training, we introduce a Modal Consistency Dynamic Balancing (MCDB) strategy, which quantifies the relative influence of each modality and adaptively adjusts the information weights in the shared representation. This balanced allocation mechanism prevents any single modality from dominating, ensuring that all modalities contribute effectively. Furthermore, to prevent the decline in model generalization ability caused by balancing modal contributions, we designed a semantic noise intervention mechanism, namely stochastic perturbation regularization, inspired by the stochastic resonance phenomenon in the visual cortex \cite{mcdonnell2011benefits}. Research on this phenomenon shows that an appropriate amount of intrinsic neural noise can amplify weak signals and improve perceptual robustness. Based on this, we inject controllable Gaussian noise into the shared multimodal representation during the training phase, forcing the network to maintain its discrimination ability under noise interference, thereby learning a more robust cross-modal semantic mapping.

In summary, our contributions are as follows:
\begin{itemize}

\item Inspired by the ability of the brain to decode representations, we propose a novel framework that can leverage textual, visual, and EEG representations to mutually enhance each other, which can improve the decoding ability of EEG.


\item For further leverage pre-trained visual and textual representation, we introduce an adapter, which aims to reduce the instability of high dimensionality and preserve semantic information.

\item To alleviate optimization imbalance, we introduce a Modal Consistency Dynamic Balancing (MCDB) strategy that quantifies the contributions of each modality and reinforces the synergy between EEG representations and multimodal integration. We further propose a Stochastic Perturbation Regularization (SPR) term that incorporates dynamic Gaussian noise during optimization to improve generalizability of the shared representation.

\end{itemize}

\section{Related Work}
\subsection{Decoding Visual Information from Brain Signals}
The neural networks of the human brain are studied using neuroimaging techniques such as functional magnetic resonance imaging (fMRI), functional near-infrared spectroscopy (fNIRS), invasive electrocorticography (iEEG), and noninvasive electroencephalography (EEG). Although fMRI provides high spatial resolution through blood oxygen level dependent (BOLD) signals, it is limited by low temporal resolution and high costs. fNIRS offers a portable and cost-effective alternative by detecting superficial changes in blood flow, but lacks temporal precision and depth. iEEG delivers both high temporal and spatial resolution via implanted electrodes, though its invasiveness restricts widespread application. In contrast, EEG is favored for its high temporal resolution, non-invasiveness, and portability, despite challenges like low signal-to-noise ratio and signal non-stationarity, which complicate semantic information extraction from images.

Early EEG research employed linear supervised methods such as Canonical Correlation Analysis (CCA) and Linear Discriminant Analysis (LDA), suitable for low-dimensional data but inadequate for high-dimensional, multi-channel signals. Techniques like Common Spatial Patterns (CSP) and time-frequency feature extraction enhanced signal discrimination related to visual stimuli. The rise of deep learning introduced end-to-end Convolutional Neural Networks (CNNs) for raw EEG signal classification, improving performance and generalization through transfer learning. However, these methods are constrained by limited labeled datasets, affecting zero-shot classification. To address data scarcity, self-supervised learning, particularly contrastive learning frameworks, has been adopted to capture EEG signals' spatio-temporal dependencies, although generalizing across modalities and complex data remains challenging.

\subsection{Multimodal Contrastive Learning} 
Real-world information is inherently multimodal, including text, images, audio, and video, which together enable comprehensive understanding \cite{meyer2020improving,miao2023ds,zhang2024mitigating,zhang2024bayesian}. Integrating multimodal data and mining their semantic correlations are key challenges in computer science. Contrastive learning has been extended to align features across modalities, capturing semantic consistency for universal representation learning \cite{yuan2021multimodal, nakada2023understanding, bansal2023cleanclip, miao2025laser,miao2025towards}. This approach enhances model generalization in low-label scenarios and cross-modal adaptability. A prominent example is CLIP \cite{radford2021learning}, which maps image and text features into a shared space, optimizing similarity for matching pairs and minimizing it for non-matching ones to capture semantic relationships. Building on CLIP, studies have expanded its framework: Wu et al. \cite{wu2022wav2clip} integrated audio without additional visual model training, reducing computational overhead; Lin et al. \cite{lin2023multimodality} developed a cross-modal few-shot learning framework using features from different modalities to enhance unimodal tasks; Xue et al. \cite{xue2024ulip} aligned 3D point clouds, 2D images, and language descriptions for large-scale trimodal data alignment without manual annotation.

However, most research focuses on visual and linguistic modalities, with limited integration of time-series signals like EEG with images. Ye et al. \cite{ye2022see} introduced an EEG-image contrastive learning method for reconstructing image representations from visual stimuli, but it relies on high-quality EEG encoders and lacks generalization in complex tasks. Song et al. \cite{song2023decoding} enhanced EEG encoders with self-attention and graph attention modules to better model temporal, spatial, and spectral features, improving zero-shot classification. Chen et al. \cite{chen2024mind} employed regularized contrastive learning to enhance EEG-image alignment by suppressing noise and optimizing cross-modal embeddings. Bai et al. \cite{bai2023dreamdiffusion} incorporated text representations during EEG-image alignment, further refining semantic accuracy. In addition, Wang et al. \cite{wang2024mindbridge} combined SoftCLIP+MSE loss with recurrent fMRI reconstruction in a unified CLIP space to achieve single-model cross-subject brain decoding, verifying the versatility of contrastive learning in brain-vision alignment. Bao et al.\cite{bao2025wills} refreshes the SOTA on NSD classification, retrieval, and reconstruction tasks through anatomical template alignment and sparse MoBE adapter, and uses semantically driven contrastive loss. Gong et al. \cite{gong2025mindtuner} proposed MindTuner, which injects Skip-LoRA into the encoder and uses the image as the pivot for fMRI-image-text trimodal alignment, achieving high-fidelity cross-subject reconstruction and retrieval with only 1 h of fMRI data.

\subsection{Disparities in Modality Contributions} 
multimodal learning can significantly enhance task performance by leveraging multiple modalities. However, differences between modalities often lead to varying convergence rates, resulting in the modality dominance effect, where models rely more on stronger modalities, diminishing weaker ones and limiting overall generalization \cite{wang2020makes}.

To mitigate dominant modality bias, various strategies have been developed. One approach involves specialized fusion techniques for efficient integration. Brockschmidt et al. \cite{brockschmidt2020gnn} introduced dynamic affine transformation functions that adjust input feature weights based on target node information, enhancing weaker modalities. Du et al. \cite{du2021improving} combined unimodal pre-training with feature distillation to transfer high-quality unimodal representations to multimodal networks, strengthening weaker modalities. Similarly, Han et al. \cite{han2021improving} maximized mutual information between modalities and the fused output, effectively filtering noise and enhancing shared features. However, these methods often increase training complexity and computational costs due to additional neural modules.

Therefore, people have begun to turn to methods that balance the optimization process by dynamically adjusting the modality weights \cite{peng2022balanced, li2023boosting}. For example, Lin et al. \cite{lin2024suppress} proposed the ReGrad strategy, which monitors cross-modal gradient conflicts and convergence speeds, adaptively scales fast modal gradients and amplifies the gradients of slower learning modalities, effectively alleviating the bias of the dominant modality. Wei et al. \cite{wei2024enhancing} proposed the Sample-level Modality Valuation method, which evaluates the contribution of each modality at the sample level based on the Shapley value, and probabilistically resamples the low-contribution modality to improve its discrimination ability in a targeted manner, significantly enhancing cross-modal collaborative performance in datasets such as MM-Debiased. These methods use adaptive gradients or sampling modulation to change the optimization strength of each modality, speed up the learning speed of the weaker modality, and ensure that it receives enough attention during training, thereby effectively balancing the optimization process of the multimodal model.

Building on these insights, this work proposes a novel optimization-based regulatory method that dynamically adjusts the optimization intensity for each modality by real-time monitoring of the contribution differences between images and text during EEG feature alignment, without introducing additional neural modules. In complex multimodal contrastive learning tasks, this method significantly improves performance, mitigating feature bias and enhancing model generalization.

\section{Method}
To achieve object recognition by learning image representations from EEG signals, we propose a novel framework, HMAVD. We first leverage textual representations to enforce semantic consistency between EEG signals and visual content in Section \ref{sec:3.1}. Then, we introduce an adapter module to stabilize high-dimensional representation and improve cross-modal feature fusion in Section \ref{sec:3.2}. Finally, we describe the dual optimization collaboration mechanism (DOCM) in Section \ref{sec:3.3}.

\begin{figure*}[htbp]
    \centering   
    \includegraphics[width=0.8\textwidth]{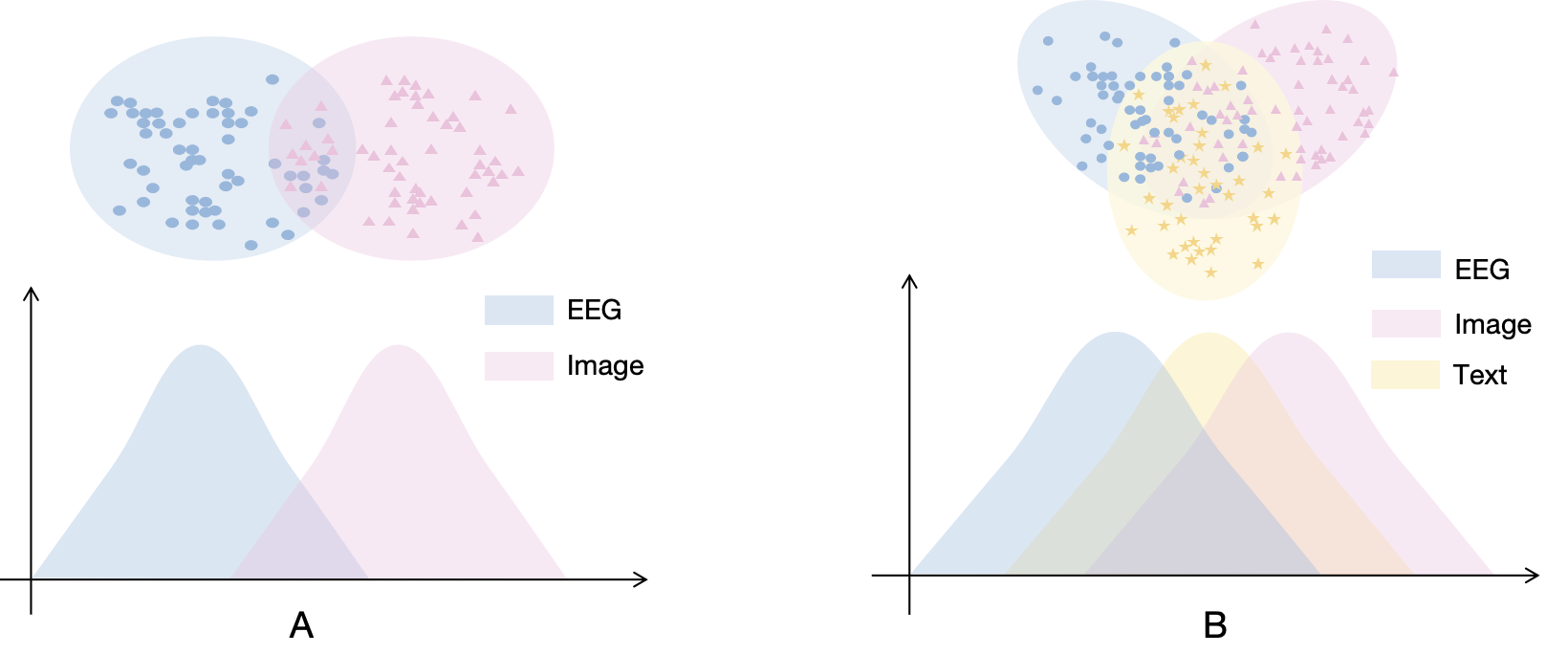}
    \caption{(A) EEG and image feature distribution in a shared space. The distribution is sparse. (B) EEG, image and text feature distribution in a shared space. After the text feature added, the distribution becomes tighter.}
    \label{fig:text}
\end{figure*}

\subsection{Text Modality for Semantic Bridging}\label{sec:3.1}
EEG signals have high temporal resolution and dynamic characteristics, however, their semantic expression ability is limited and they are susceptible to noise interference and non-stationarity. In contrast, the image modality has stronger semantic expression ability. This difference complicates the direct alignment of EEG representation \( f_{\text{eeg}} \) and image representation \( f_{\text{img}} \) and may lead to feature sparsity and optimization instability in the shared embedding space (as shown in Figure \ref{fig:text} (A)). To address this problem, we introduce the text modality \( f_{\text{text}} \) as a bridge to enhance the semantic correspondence between EEG signals and visual content. Specifically, \( f_{\text{img}} \), \( f_{\text{eeg}} \), and \( f_{\text{text}} \) are mapped to a shared multimodal space \( \mathcal{S} \), which is defined as follows:
\begin{equation}
\mathcal{S} = \left\{ \phi_{\text{eeg}}(f_{\text{eeg}}), \phi_{\text{img}}(f_{\text{img}}), \phi_{\text{text}}(f_{\text{text}}) \right\}
\end{equation}
Where, \( \phi_{\text{eeg}} \), \( \phi_{\text{img}} \), and \( \phi_{\text{text}} \) represent the mapping functions of EEG, image, and text features, respectively. In the shared space \( \mathcal{S} \), there is a significant difference in the distribution of \( f_{\text{img}} \) and \( f_{\text{eeg}} \). By introducing the explicit semantic labels provided by \( f_{\text{text}} \), \( f_{\text{img}} \) and \( f_{\text{eeg}} \) of the same category can be more closely aligned with the corresponding text representations in the shared multimodal space, thereby improving the feature distribution in the shared space (as shown in Figure \ref{fig:text} (B)).

\paragraph{Comprehensive Loss Function}
To achieve semantic alignment of EEG, image, and text representations in a shared mapping space, we design the following comprehensive loss function to optimize the alignment between modalities:
\begin{equation}
L = \frac{1}{5} \left( L_{\text{CE}}^{\text{eeg-img}} + L_{\text{CE}}^{\text{img-eeg}} + L_{\text{CE}}^{\text{eeg-text}} + L_{\text{CE}}^{\text{text-eeg}} + L_{\text{cos}} \right)    
\label{eq:loss function}
\end{equation}
Here, the cross-entropy loss function \( L_{\text{CE}} \) is defined as:
\begin{equation}
L_{\text{CE}} = -\frac{1}{B} \sum_{i=1}^B \log \frac{\exp\left(\frac{\mathbf{f}_i^{(\text{eeg})} \cdot \mathbf{f}_i^{(m)}}{\tau}\right)}{\sum_{j=1}^B \exp\left(\frac{\mathbf{f}_i^{(\text{eeg})} \cdot \mathbf{f}_j^{(m)}}{\tau}\right)}
\end{equation}
where \( m \) represents the modality pair (such as \( \text{img} \) or \( \text{text} \)), \( B \) is the batch size, \( \mathbf{f}_i^{(\text{eeg})} \) and \( \mathbf{f}_j^{(m)} \) are the feature vectors of EEG and modality \( m \) respectively, and \( \tau \) is the temperature parameter used to control the scale of similarity. Using this formula, we define the following four cross-entropy loss terms $
L_{\text{CE}}^{\text{eeg-img}}, L_{\text{CE}}^{\text{img-eeg}}, L_{\text{CE}}^{\text{eeg-text}}, L_{\text{CE}}^{\text{text-eeg}}
$.
These loss functions aim to maximize the similarity between each pair of modalities, thereby minimizing the corresponding semantic distances.

Additionally, the cosine similarity loss \( L_{\text{cos}} \) aims to maximize the cosine similarity between image features and their corresponding text features, which is defined as:
\begin{equation}
L_{\text{cos}} = 1 - \frac{1}{B} \sum_{i=1}^B \frac{\langle f_{\text{img}, i}, f_{\text{text}, i} \rangle}{\|f_{\text{img}, i}\| \|f_{\text{text}, i}\|}
\end{equation}
where, $\langle \mathbf{f}_{\text{img}, i}, \mathbf{f}_{\text{text}, i} \rangle$ represents the inner product of image features and text features, which is used to measure the similarity between the two. $\|\mathbf{f}_{\text{img}, i}\|$ and $\|\mathbf{f}_{\text{text}, i}\|$ are the $L_2$ norms of image and text features, respectively, which are used to normalize the feature vector so that the similarity measure only depends on the direction of the feature vector.

By optimizing the above loss terms, we establish closer semantic connections in the shared embedding space, allowing EEG, image, and text modalities to complement each other, reducing semantic bias between modalities and enhancing the alignment of multimodal features.

\begin{figure*}[htbp]
    \centering       \includegraphics[width=0.8\textwidth]{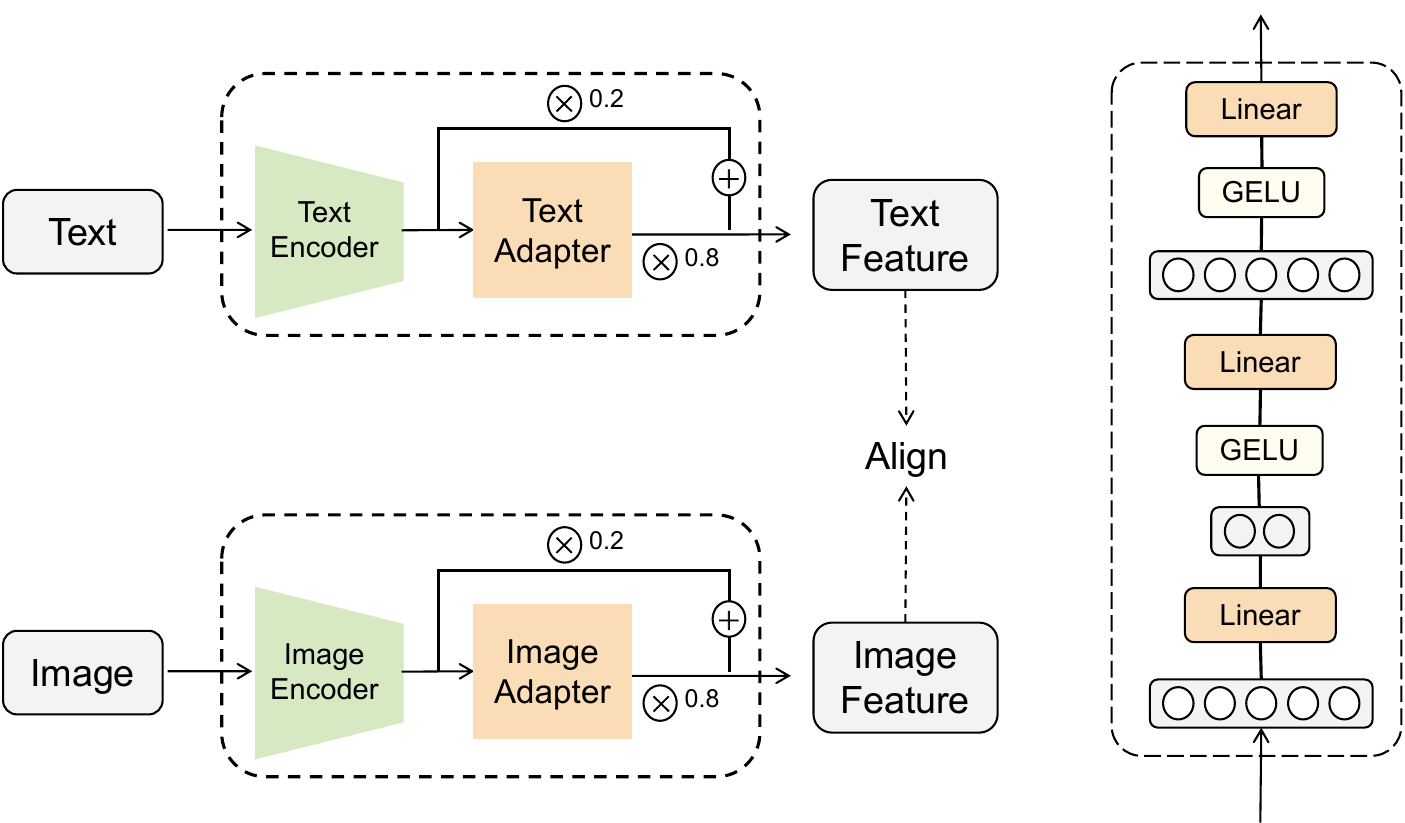}
    \caption{Structure of the adapter layer (right). The adapter layer consists of two linear layers, two activation functions and a dropout layer connection.}
    \label{fig:Adapter}
\end{figure*}

\subsection{Adapter Module}\label{sec:3.2}
\paragraph{Module Design}
Introducing the text modality significantly enhances the semantic consistency between EEG and image modalities. However, the high-dimensional nature and complex semantic structure of the text modality pose new challenges for cross-modal learning. Directly using text features may lead to unstable feature distributions within the shared representation space, thereby affecting alignment performance. Moreover, the differences in feature distributions between the image modality and the EEG and text modalities may also reduce alignment efficiency. To address this, we designed an adapter module based on a bottleneck structure, as shown in Fig \ref{fig:Adapter}. This module captures global dependencies through an attention mechanism to enhance semantic representation, while the bottleneck layer reduces dimensionality and noise, preserving core semantic information.

The mathematical definition of the adapter module is as follows: for the image modality, the adapter is represented as
\begin{equation}
A_v(x) = x + \alpha \cdot W_2(\sigma(W_1 x))
\end{equation}
where $\alpha \in (0,1)$ is a fixed residual scaling factor, \( W_1 \in \mathbb{R}^{d \times r} \), \( W_2 \in \mathbb{R}^{r \times d} \), \( \sigma \) denotes the activation function (such as ReLU), and \( r \) is the compression ratio used to reduce dimensions while retaining core features. For the text modality, the adapter is designed as
\begin{equation}
A_t(x) = x + \alpha \cdot W_2(\sigma(W_1 x + b_1)) + b_2
\end{equation}
where $b_1$ and $b_2$ are bias terms used to enhance the capability of nonlinear transformations. Given that text features have significantly higher dimensionality than images, we use compression ratios of $r = 16$ for text and $r = 8$ for images. The specific compression ratios for text and images are empirically determined to strike an effective balance between reducing computational complexity and retaining sufficient semantic information. A larger $r$ helps suppress the inherent high-dimensional complexity of the text modality, while a smaller $r$ ensures that key visual features are not lost while maintaining computational efficiency.

To improve the semantic consistency between the optimized features and the original features, we introduce cosine similarity as the optimization objective. The loss function of the adapter module is defined as:
\begin{equation}
L_r = \cos \langle A_t(F_{\text{text}}), F_{\text{text}} \rangle
\end{equation}
where \( F_{\text{text}} \) is the original text feature, and \( A_t(F_{\text{text}}) \) is the adapter-optimized text feature. This loss function effectively suppresses noise interference while preserving semantic information by maximizing the cosine similarity between the adapter output and the original features.

Furthermore, to balance the contribution between the optimized features and the original features, we fuse the two through residual connections to generate new text modality features:
\begin{equation}
F_{\text{text}}^{\text{new}} = \beta A_t(F_{\text{text}}) + (1 - \beta) F_{\text{text}}
\end{equation}
where \( \beta \) is the weight parameter used to control the proportion of optimized features and original features.

\begin{figure}[ht]
    \centering
    \includegraphics[width=0.8\textwidth]{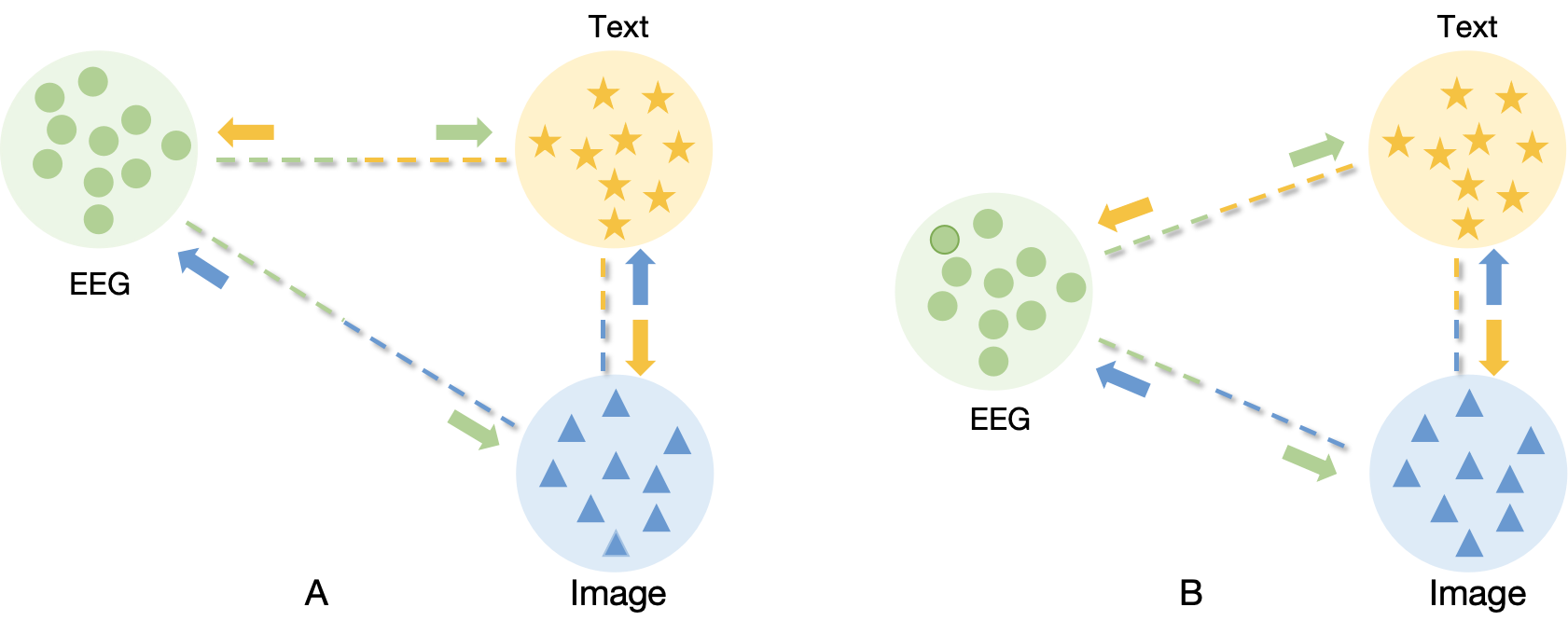}
    \caption{Effects of dynamic gradient modulation on EEG optimization: pre-modulation optimization is dominated by a single modality, limiting the potential of another modality (A); post-modulation optimization is influenced by a balance of the two modalities, taking full advantage of the multimodal potential (B).}
    \label{fig:contribution balance}
\end{figure}

\subsection{Dual Optimization Collaborative Mechanism }\label{sec:3.3}

In a shared multimodal space, image modality features are stable, while text modality features are semantically complex and variable. This imbalance during feature alignment causes EEG features to be unevenly influenced by image and text modalities, weakening their unique expressive capacity and reducing overall multimodal collaboration effectiveness. To address this issue, this work proposes a Dual Optimization Collaborative Mechanism (DOCM), whose core components are Modality Consistency Dynamic Balance (MCDB) and Stochastic Perturbation Regularization (SPR). MCDB dynamically adjusts modality weights to mitigate the over-dominance of a single modality over EEG features, while SPR introduces controllable noise to enhance the robustness and generalization ability of the alignment process.

\paragraph{Modality Consistency Dynamic Balance}

In joint learning of EEG, image, and text modalities, distribution differences in modality features can lead to dominant modalities controlling the shared embedding space, suppressing the information expression of other modalities (as shown in Fig \ref{fig:contribution balance} (A)). This modality imbalance weakens the robustness and performance of multimodal learning.

To mitigate the dominance effect of certain modalities, we design MCDB to dynamically adjust the contributions of each modality to the shared space, ensuring balance between modalities during the alignment process (as shown in Fig \ref{fig:contribution balance} (B)). Let the training dataset be \( \mathcal{D} = \{(\mathbf{x}_i, y_i)\}_{i=1}^N \), where \( \mathbf{x}_i = (\mathbf{x}_i^{\text{eeg}}, \mathbf{x}_i^{\text{img}}, \mathbf{x}_i^{\text{text}}) \) contains EEG, image, and text modalities, and \( y_i \) is the corresponding label. The feature of modality \( m \in \{\text{eeg}, \text{img}, \text{text}\} \) is extracted by its encoder \( \psi^{(m)} \) and represented as:
\begin{equation}
\mathbf{f}_i^{(m)} = \psi^{(m)}(\mathbf{x}_i^{(m)}; \boldsymbol{\theta}^{(m)})
\end{equation}

Where $\theta^{(m)}$ denotes the weight of the encoder \(\psi^{(m)}\), which are optimized during training. Within a batch, define the batch feature matrix of modality \( m \) as \( \mathbf{F}^{(m)} \in \mathbb{R}^{B \times d} \), where \( B \) is the batch size, and \( \mathbf{F}^{(m)} = [\mathbf{f}_1^{(m)}, \mathbf{f}_2^{(m)}, \dots, \mathbf{f}_B^{(m)}]^T \). To quantify the relative contribution between modalities, we compute the similarity matrix between modalities:
\begin{equation}
\mathbf{S}^{(m_1, m_2)} = \text{softmax}\left(\frac{\mathbf{F}^{(m_1)} (\mathbf{F}^{(m_2)})^T}{\tau}\right)
\end{equation}
where \( \tau > 0 \) is the temperature parameter that controls the smoothness of the distribution. By summing the columns of the similarity matrix, we obtain the total contribution of modality \( m \) to the EEG features:
\begin{equation}
\mathbf{s}^{(\text{eeg}, m)} = \sum_{j=1}^B \mathbf{S}^{(\text{eeg}, m)}_{:, j}
\end{equation}

Based on the contribution values, we define the Modality Imbalance Rate:
\begin{equation}
\rho^{(m)} = \frac{\|\mathbf{s}^{(\text{eeg}, m)}\|_1}{\sum_{n \neq m} \|\mathbf{s}^{(\text{eeg}, n)}\|_1 + \epsilon}
\end{equation}
where \( \|\cdot\|_1 \) denotes the \( \ell_1 \)-norm and \( \epsilon > 0 \) is a smoothing factor to prevent division by zero. When \( \rho^{(m)} > 1 \), it indicates that modality \( m \) has an excessive contribution to the shared space.

To dynamically adjust modality contributions, we design an adaptive weight function \( \kappa^{(m)} \):
\begin{equation}
\kappa^{(m)} = 
\begin{cases} 
1 - \tanh(\gamma \cdot (\rho^{(m)} - 1)) & \text{if } \rho^{(m)} > 1 \\ 
1 & \text{if } \rho^{(m)} \leq 1
\end{cases}
\end{equation}
where \( \gamma > 0 \) controls the sensitivity of the adjustment. Finally, we incorporate \( \kappa^{(m)} \) into the optimization objective so that the gradient updates of each modality are dynamically regulated, thereby achieving modality balance.

Based on the joint loss function \( L \) (Equation \eqref{eq:loss function}), MCDB integrates \( \kappa^{(m)} \) into the optimization process, making the gradient update formula for modality \( m \):
\begin{equation}
\nabla_{\boldsymbol{\theta}^{(m)}} L = \kappa^{(m)} \cdot \frac{\partial L}{\partial \mathbf{f}_i^{(m)}} \cdot \frac{\partial \mathbf{f}_i^{(m)}}{\partial \boldsymbol{\theta}^{(m)}}
\end{equation}
Specifically, the gradient update process of MCDB is as follows:
\begin{equation}
\boldsymbol{\theta}_{t+1}^{(m)} = \boldsymbol{\theta}_{t}^{(m)} - \eta \cdot \kappa^{(m)} \cdot \nabla_{\boldsymbol{\theta}^{(m)}} L
\label{eq:gradient descent}
\end{equation}
where \( \eta \) is the learning rate. When \( \rho^{(m)} > 1 \), \( \kappa^{(m)} < 1 \), and the gradient updates of the dominant modality are suppressed; whereas when \( \rho^{(m)} \leq 1 \), \( \kappa^{(m)} = 1 \), ensuring that the contributions of weaker modalities are fully retained.

\paragraph{Stochastic Perturbation Regularization}

In multimodal tasks, EEG features, due to their inherent high noise characteristics and low resolution, are often dominated by more stable modalities during optimization. This phenomenon may lead to local optima in multimodal alignment. Traditional optimizers partially alleviate gradient oscillations through momentum but fail to resolve the issue of modality contribution imbalance. Therefore, we propose SPR, which introduces dynamic noise to further enhance the exploration capability of the optimizer.

In Equation \eqref{eq:gradient descent}, the dynamic adjustment of modality weights \( \kappa^{(m)} \) effectively alleviates the imbalance of modality contributions. However, in complex multimodal tasks, this formula alone is insufficient to escape local optima. Therefore, SPR further supplements the optimization process by incorporating random noise.

To enhance the exploration capability of the optimization process, SPR introduces a random noise term \( h(\theta_t) \) into the gradient update, extending the formula as follows:
\begin{equation}
    \theta_{t+1}^{(m)} = \theta_t^{(m)} - \eta \cdot \left( \kappa^{(m)} \cdot \nabla_{\theta^{(m)}} L + h(\theta_t) \right)
\end{equation}
where $h(\theta_t) \sim \mathcal{N}(0, \sigma_t^2 I)$ is a multidimensional Gaussian noise with zero mean and variance \( \sigma_t^2 \), and \( \sigma_t \) is the dynamically controlled noise intensity.

The introduction of the noise term \( h(\theta_t) \) adds randomness to the optimization path, enabling the model to escape local optima and explore better parameter spaces.

To balance the exploration capability and convergence of the optimization process, the noise intensity \( \sigma_t \) is dynamically adjusted with training iterations, as shown below:
\begin{equation}
    \sigma_t = \sigma_{\text{max}} \cdot \left( 1 - \frac{t}{T} \right)^\beta
\end{equation}
where $\sigma_{\text{max}}$ is the initial noise intensity, $t$ is the current training iteration,
 $T$ is the total training iterations, $\beta$ is the noise decay factor.

To further improve optimization efficiency, SPR is integrated into momentum-based optimizers (Adam). The update rules for the momentum terms are as follows:
\begin{equation}
    m_t = \beta_1 m_{t-1} + (1 - \beta_1) \cdot \left( \kappa^{(m)} \cdot \nabla_{\theta^{(m)}} L + h(\theta_t) \right)
\end{equation}
\begin{equation}
    v_t = \beta_2 v_{t-1} + (1 - \beta_2) \cdot \left( \kappa^{(m)} \cdot \nabla_{\theta^{(m)}} L + h(\theta_t) \right)^2
\end{equation}
where $m_t$ is the first-order momentum term for smoothing gradient updates, $v_t$ is the second-order momentum term for measuring the weighted average of gradient squares, $\beta_1, \beta_2$ are momentum decay coefficients.

The final parameter update formula is:
\begin{equation}
    \theta_{t+1}^{(m)} = \theta_t^{(m)} - \eta \cdot \frac{m_t}{\sqrt{v_t} + \epsilon}
\end{equation}
where \( \epsilon \) is a small constant to prevent numerical instability. By incorporating the noise term \( h(\theta_t) \), the optimizer retains the stability of momentum-based optimization while enhancing its exploration capability.

\section{Experiment}
\subsection{Dataset and Preprocessing}
This work employs the extended ThingsEEG framework to create a multimodal dataset consisting of 16,740 natural images, their corresponding text labels, and EEG signals. EEG data were recorded from 10 participants using a rapid continuous presentation paradigm to enhance efficiency and reduce interference. The training set comprises 1,654 categories, each containing 10 images repeated four times, while the test set includes 200 categories, each with one image repeated 80 times. Images were presented in a pseudo-random order, each displayed for 100 milliseconds with a 100-millisecond blank screen interval, and target images served as calibration references. EEG signals were collected through 63 channels, focusing on 17 channels related to visual functions, with a frequency range of 0.1–100 Hz and a sampling rate of 1000 Hz.

For extracting visual-related neural features, the EEG signals were first segmented into 1000-millisecond intervals starting from stimulus onset, and baseline correction was performed using the mean of the preceding 200 milliseconds to reduce drift. Subsequently, the signals were downsampled to 250 Hz to decrease computational load while maintaining temporal resolution. Next, multivariate noise normalization was applied to the EEG signals in the training set. Finally, the EEG recordings from all repeated trials for each image were averaged to obtain more stable neural response features.

\subsection{Experiment Details}
Our method is implemented in PyTorch and trained on a 4090 GPU.  For each run, we randomly sample $740$ trials from the training data to construct a validation set and keep the checkpoint with the lowest validation loss for subsequent evaluation. We adopt the default temperature of \textsc{CLIP}, setting $\tau = 0.07$ (logit scale initialized to $\ln(1/\tau) \approx 2.659$).   The modulation coefficient is fixed at $\alpha = 0.7$.   Training employs the Adam optimizer \emph{without} weight decay, with momentum parameters $(\beta_1, \beta_2) = (0.5, 0.999)$.  The initial learning rate is $2 \times 10^{-4}$.  Mini-batch sizes are $1000$ for training and $400$ for testing.  The network is trained for $200$ epochs, after which we report performance on the held-out test set.

\begin{table}[htbp!]
  \centering
  \caption{Comparison of computation time for our and NICE.}
  \resizebox{0.42\linewidth}{!}{          
    \begin{tabular}{l|cc}
      \toprule
      Method & Time (min) & Relative time \\
      \midrule
      our  & 7.35 & 1.59 \\
      NICE & 4.63 & 1.00 \\
      \bottomrule
    \end{tabular}
  }
  \label{tab:runtime_compare}
\end{table}

\subsection{computation Efficiency}
We benchmarked the runtime efficiency of Subject 1 and directly compared our method with NICE. The results are shown in Table \ref{tab:runtime_compare}. Under the same hardware and data settings, the computational efficiency of our method is reduced due to the additional calculation of dynamic modal weight adjustment and random perturbation regularization required by our MCDB and SPR mechanisms. However, the core goal of our method is to improve the robustness and generalization performance of multimodal feature fusion, rather than simply pursuing computational efficiency in the inference phase.

\begin{table*}[htbp!]
\caption{\textbf{Performance Comparison on EEG Decoding for Object Recognition: Top-1 / Top-5}}
\renewcommand{\arraystretch}{1.5} 
\setlength{\tabcolsep}{4pt} 
\centering
\scriptsize 
\resizebox{\textwidth}{!}{
\begin{tabular}{l|cccccccccccccccccccccccccccccccccccccccccccccccccccccccccccccccccccccccccccccccccccccccccccccccccccccccccccccccccccccccccccccccccccccc|c}
\toprule
\textbf{Method} & \textbf{Sub1} & \textbf{Sub2} & \textbf{Sub3} & \textbf{Sub4} & \textbf{Sub5} & \textbf{Sub6} & \textbf{Sub7} & \textbf{Sub8} &  \textbf{Sub9} & \textbf{Sub10} &\textbf{Avg.} \\
\midrule
BraVL \cite{du2023decoding} & 6.1/17.9 & 4.9/14.9 & 5.6/17.4 & 5.0/15.1 & 4.0/13.4 & 6.0/18.2 & 6.5/20.4 & 8.8/23.7 & 4.3/14.0 & 7.0/19.7 & 5.8/17.5 \\
NICE  \cite{song2023decoding} & 12.3/36.6 & 10.4/33.9 & 13.1/39.0 & 16.4/47.0 & 8.0/26.9 & 14.1/40.6 & 15.2/42.1 & 20.0/49.9 & 13.3/37.1 & 14.9/41.9 & 13.8/ 39.5 \\
HMAVD-17 & 9.5/39.0 & 10.5/37.0 & 15.5/40.0 & 17.0/42.5 & 7.0/32.5 & 15.0/46.5 & 13.0/43.5 & 21.0/55.5 & 16.0/43.0 & 11.5/43.0 & 13.6/42.3 \\
HMAVD-63 & \cellcolor[HTML]{FED7DC}14.5/39.0 & \cellcolor[HTML]{FED7DC}14.5/42.5 & \cellcolor[HTML]{FED7DC}13.5/42.5 & \cellcolor[HTML]{FED7DC}17.5/51.0 & \cellcolor[HTML]{FED7DC}11.0/32.5 & \cellcolor[HTML]{FED7DC}17.0/45.0 & \cellcolor[HTML]{FED7DC}16.5/47.0 & \cellcolor[HTML]{FED7DC}21.5/55.5 & \cellcolor[HTML]{FED7DC}15.5/43.0 & \cellcolor[HTML]{FED7DC}16.5/43.5 & \cellcolor[HTML]{FED7DC}15.8/44.2\\
\hline
NICE-SA  \cite{song2023decoding} & 13.3/40.2 & 12.1/36.1 & 15.3/36.9 & 15.9/49.0 & 9.8/34.4 & 14.2/42.4 & 17.9/43.6 & 18.2/50.2 & 14.4/38.7 & 16.0/42.8 & 14.7/41.7 \\
HMAVD-63-SA & 14.5/39.5 & 14.0/35.0 & 17.0/39.5 & 18.5/47.5 & 11.5/36.5 & 14.5/46.5 & 17.0/44.5 & 21.0/53.0 & 17.0/44.0 & 16.0/43.0 & 16.1/42.9 \\
\hline
NICE-GA  \cite{song2023decoding} & 15.2/40.1 & 13.9/40.1 & 14.7/42.7 & 17.6/48.9 & 9.0/29.7 & 16.4/44.4 & 14.9/43.1 & 20.3/52.1 & 14.1/39.7 & 19.6/46.7 & 15.6/42.8 \\
HMAVD-63-GA & 16.0/39.5 & 16.0/42.0 & 18.0/46.0 & 18.0/48.0 & 16.5/33.5 & 14.0/41.5 & 17.0/47.0 & 19.0/55.5 & 16.0/43.5 & 15.0/50.5 & 16.5/44.7 \\
\bottomrule
\end{tabular}
}
\label{tab:results}
\end{table*}

\subsection{Comparison with State-of-the-Art Methods}
\paragraph{Quantitative Results}
We conducted a comprehensive evaluation of the proposed framework, HMAVD, on the latest and largest image-EEG dataset, ThingsEEG, and compared its performance with the current state-of-the-art EEG image recognition frameworks, BraVL and NICE. As shown in Table \ref{tab:results}, HMAVD achieved an average Top-1 accuracy and Top-5 accuracy of $15.8\%$ and $44.2\%$, respectively, on data from ten subjects collected with $63$ channels. This represents an improvement of $10.0\%$ and $26.7\%$ over BraVL, and $2.0\%$ and $4.7\%$ over NICE, respectively.

Furthermore, we evaluated the decoding performance using data collected from $17$ channels that are closely related to visual functions. In this setting, HMAVD achieved a Top-1 accuracy of $13.6\%$ and a Top-5 accuracy of $42.3\%$. Although the Top-1 accuracy was slightly lower than that of NICE ($13.8\%$), the Top-5 accuracy surpassed NICE ($39.5\%$), and both Top-1 and Top-5 accuracies were significantly better than those of BraVL ($5.8\%$ and $17.5\%$). Compared to the results with $63$ channels, the Top-1 accuracy with $17$ channels was slightly lower, but the Top-5 accuracy showed almost no difference. This indicates that these $17$ channels capture the majority of key information related to vision while effectively reducing noise interference and the impact of non-essential information. Under limited computational resources, selecting $17$ channels can significantly reduce data processing and model training complexity while still maintaining excellent decoding performance.

Notably, under identical attention configurations, both HMAVD-63-GA and HMAVD-63-SA achieved superior Top-1 and Top-5 accuracy compared to their NICE counterparts. Specifically, HMAVD-63-GA attained 16.5\% Top-1 accuracy and 44.7\% Top-5 accuracy, surpassing NICE-GA (15.2\% and 42.8\%, respectively). Similarly, HMAVD-63-SA reached 16.1\% Top-1 accuracy and 42.9\% Top-5 accuracy, marginally outperforming NICE-SA (15.2\% and 41.7\%). These results demonstrate that the HMAVD architecture possesses enhanced representational capacity for EEG-visual feature fusion.

\begin{figure}[t]
    \centering
        \includegraphics[width=\textwidth]{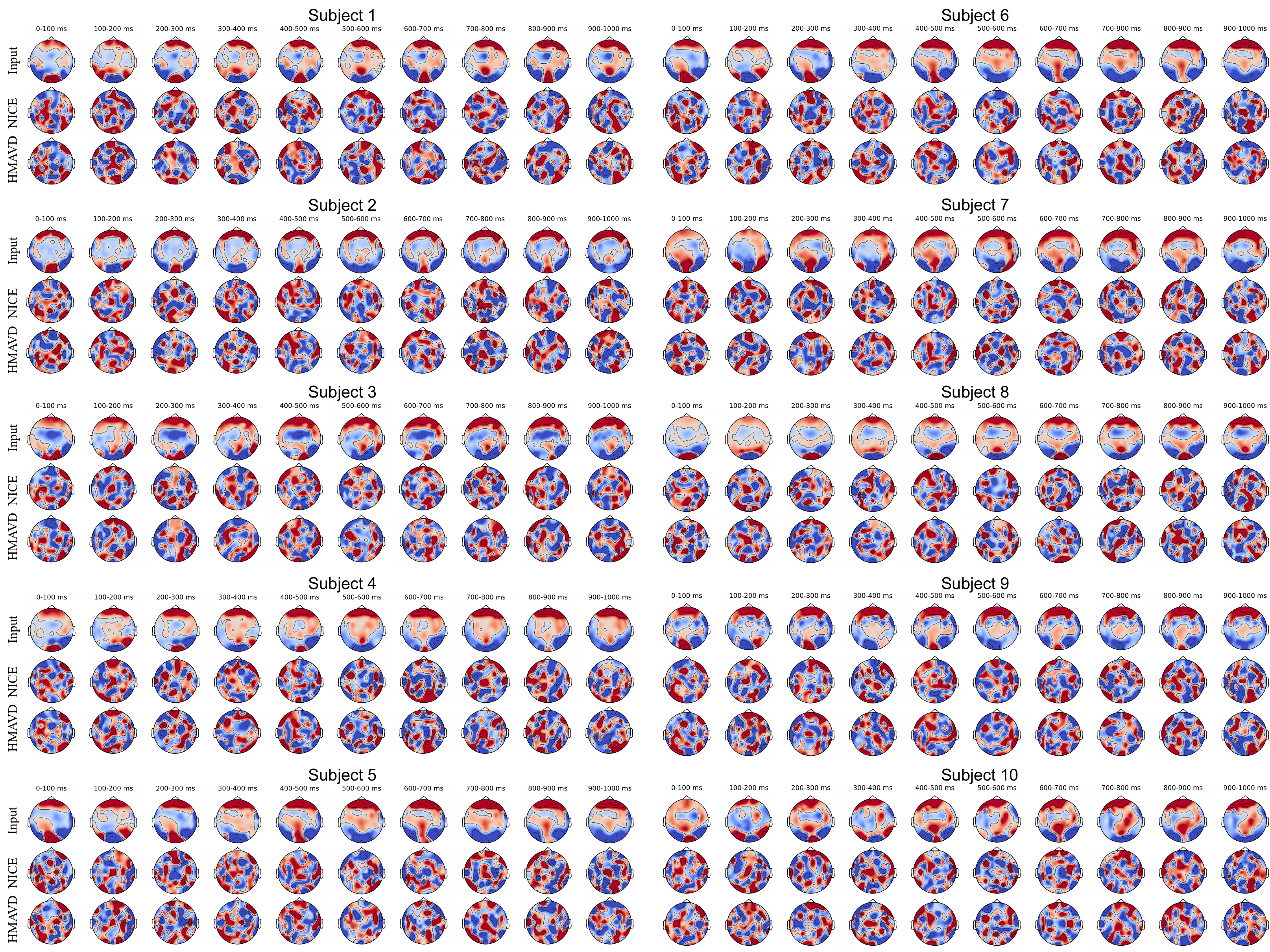}
        \caption{Dynamic Cortical Activity Topographies Averaged Over 100 ms Intervals: Initial Visual Processing in the Occipital Lobe (0–100 ms) Expands to Temporal and Parietal Regions (100–600 ms), Reflecting Feature Integration and Attentional Regulation, and Progresses to Frontal Regions in the Late Phase (600–1000 ms), Indicating Higher-Order Cognitive Processes and Task Feedback.}
    \label{fig:topo}
\end{figure}

\paragraph{Qualitative Results}
In Figure \ref{fig:topo}, we present a comparative set of topographic maps of cortical activity under three different processing conditions: raw data(Input), data processed by the NICE model(NICE), and data processed by our model(HMAVD). These maps are derived from EEG recordings of 10 subjects and depict the neural activity characteristics of the cortex with a temporal resolution of 100 ms.

In the initial 100 ms after the presentation of visual stimuli, EEG signals primarily reflect the rapid perceptual processes of the primary visual cortex (V1, V2, V3), including the extraction of low-level visual features such as color, edges, and contrast. Under the Input condition, responses were observed in the occipital region for all subjects. However, due to inherent noise in EEG signals and inter-individual variability, some subjects exhibited relatively weak occipital activation, resulting in an uneven signal distribution. NICE enhanced the activation level in the occipital region for some subjects, yet this enhancement was unstable, with signals remaining dispersed in certain cases. In contrast, HMAVD produced clearer boundaries and a more uniform activation pattern in the occipital region in multiple subjects, maintaining high consistency. This suggests that HMAVD more effectively extracts key signals from the visual cortex, reduces intersubject variability, and improves the decoding quality of low-level visual information.

Between 100 and 600 ms, as visual information propagates further within the brain, the temporal and parietal lobes begin to participate in the integration of semantic features and the modulation of spatial attention. Under the input condition, signal patterns in the temporal and parietal lobes varied considerably between subjects, and some individuals showed weak activation in specific regions. After NICE processing, the enhancement of temporal lobe signals was notable, indicating an improvement in semantic information processing for some subjects. However, significant differences in signal strength persisted in the parietal region, suggesting that NICE has limitations in regulating spatial attention allocation. In contrast, HMAVD-processed EEG signals exhibited a more coherent temporo-parietal coactivation pattern, particularly during the target recognition phase (300–600 ms), where parietal signals in several subjects showed an increasing trend. These findings indicate that HMAVD can more reliably capture cross-subject semantic integration processes while optimizing brain activity related to target recognition and attention modulation.

During the later stage from 600 to 1000 ms, the brain enters a phase of higher-level cognitive processing. Under the Input condition, activation patterns in the frontal region showed considerable inter-individual variability, with some subjects failing to exhibit clear task-related signals. Following NICE processing, frontal signals improved in some subjects, suggesting a modest optimization effect in task-feedback-related brain regions. However, overall enhancement was limited. In contrast, HMAVD generated frontal signals were not only more stable within individual subjects, but also demonstrated greater consistency between subjects, indicating greater efficacy in optimizing EEG features associated with advanced cognitive processing.

\begin{figure}[t]
    \centering
        \includegraphics[width=\textwidth]{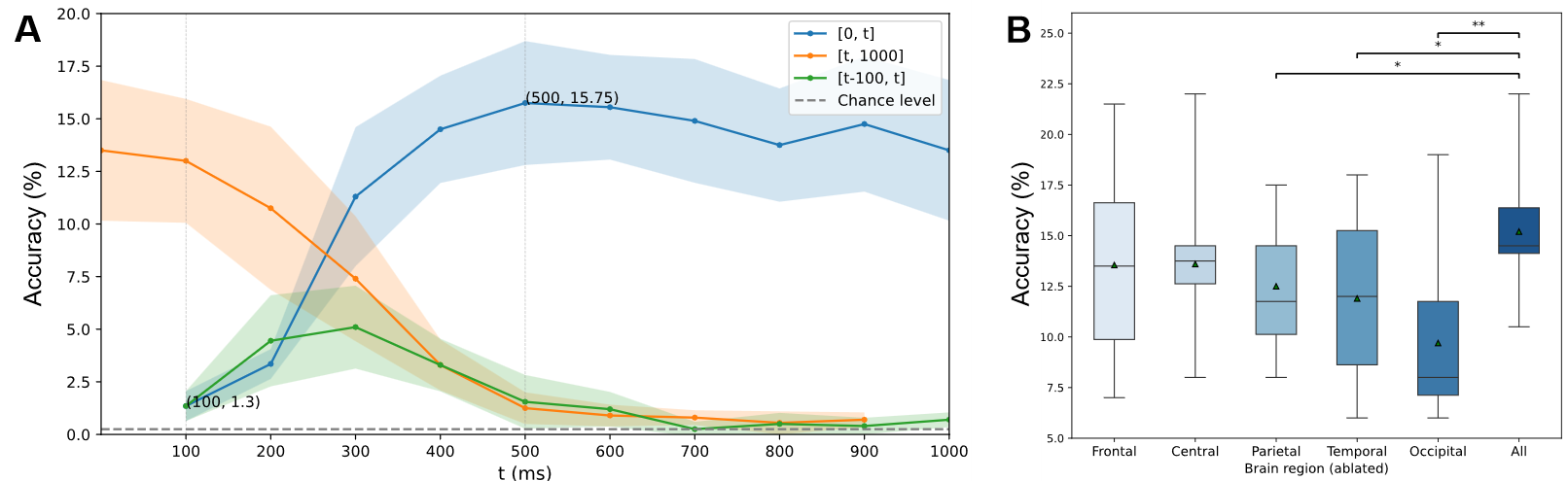}
        \caption{(A) The average accuracy of 10 subjects in different time periods, with the 100-600 ms time period being the most accurate and containing the richest visual information. (B) Ablation experiments in different brain regions. Neural activity in occipital, temporal and parietal cortex was highly predictive of visual image reconstruction.}
    \label{fig:time and spatial}
\end{figure}

\subsection{Biological Rationality Analysis}
To verify the biological plausibility of EEG signals in image classification tasks, we conducted analyses from both temporal and spatial dimensions using time window analysis and cortical ablation experiments.

In the temporal dimension analysis, we employed three methods—incremental forward, decremental backward, and segmentation—to investigate the key time intervals involved in the visual decoding process of EEG signals. As shown in Figure \ref{fig:time and spatial} (A), within the time window 0 to 100 ms, the classification accuracy based on EEG was only 1.3\%, slightly above the chance level. This finding is consistent with the early stage of visual processing, which mainly involves the extraction of basic visual features such as color, shape, and texture, without yet achieving full object recognition. As time progresses, the classification performance of the EEG signals improves significantly, reaching a peak around 500 ms. This trend aligns closely with the activation patterns observed in the temporal visual processing regions, indicating that the brain has completed object categorization and entered the semantic understanding phase. Notably, in the 600–1000 ms time window, classification performance gradually declines to near chance levels, possibly due to the involvement of higher cognitive processes such as memory retrieval and attention shifting, which may dilute the object recognition-related information contained in the EEG signals. These results suggest that EEG signals dynamically reflect the temporal characteristics of visual information processing, with changes in classification performance closely mirroring the different stages of visual cognition. 

In the spatial dimension analysis, cortical ablation experiments (Figure \ref{fig:time and spatial} (B)) were conducted to examine the contributions of various brain regions to visual processing. The experimental results indicate that removal of the occipital electrodes leads to a significant decrease of 4.6\% in the accuracy of the EEG classification. As a core region in visual processing, the occipital cortex is responsible not only for extracting primary visual features but also for transmitting information to the temporal lobe to support advanced object recognition. Eliminating EEG signals from the occipital region prevents the classification model from accessing crucial early visual information, resulting in incomplete input to higher-level regions such as the temporal lobe, thereby weakening the overall visual decoding process and significantly reducing classification accuracy. Furthermore, the removal of electrodes from the parietal and central regions (adjacent to the motor cortex) resulted in an approximate 2\% decrease in classification accuracy, underscoring the importance of these areas in visual and spatial attention mechanisms relevant to object recognition. Interestingly, the removal of temporal electrodes caused only a 1\% drop in classification accuracy, which may be attributed to the fact that EEG primarily records signals from the cortical surface, whereas the temporal lobe lies deeper within the brain and thus contributes relatively less to the recorded signal. Of particular note, the removal of frontal electrodes did not reduce classification accuracy; instead, it produced a slight increase of 0.25\%. This phenomenon might be explained by the functional characteristics of the frontal cortex, which is predominantly involved in high-level cognitive functions such as decision-making, memory, emotional regulation, and attention control—functions that have limited direct relevance to the EEG-based object classification task and may even introduce additional noise that affects classification performance.

\begin{figure}[t]
    \centering
        \includegraphics[width=\textwidth]{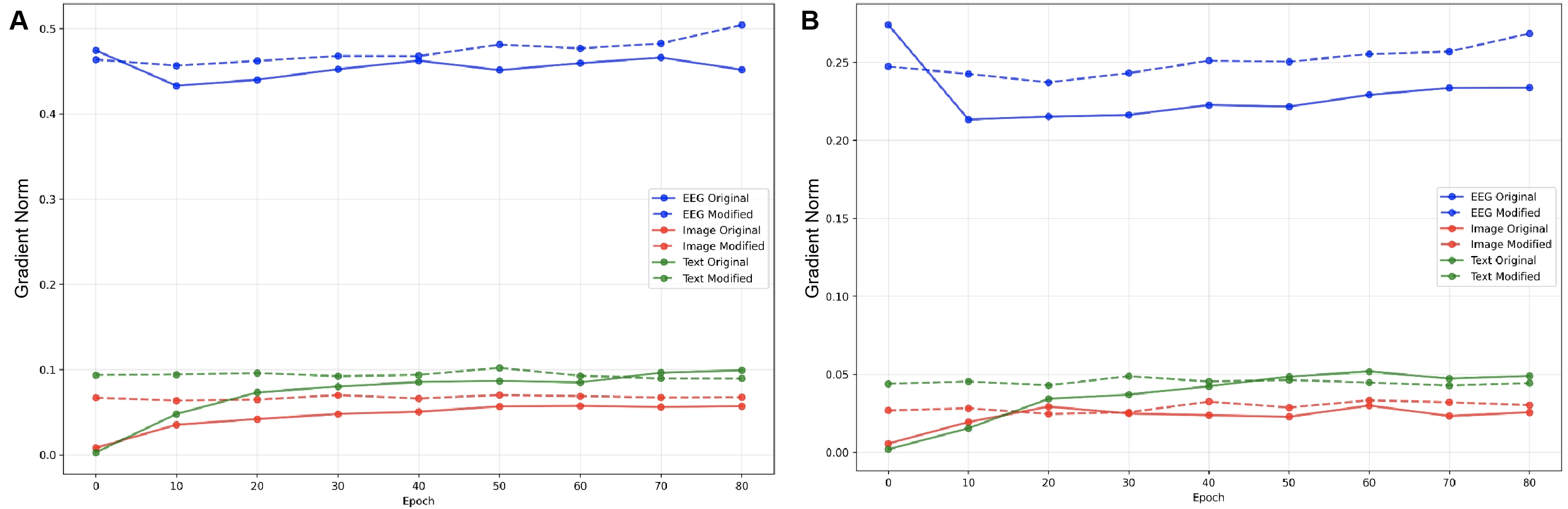}
        \caption{(A) Gradient norm trajectories for Subject 1 before and after modulation. The text modality shows dominance prior to modulation, while the modulation mechanism stabilizes the text gradients, enhances the image gradients, and improves the stability of the EEG gradients. (B) Average gradient norm trajectories across Subjects 1–10. The trends validate that the modulation mechanism effectively balances optimization resources across modalities, fostering collaborative learning and improving EEG modality performance.}
    \label{fig:visual_comparison}
\end{figure}

\subsection{Gradient Analysis}
In the optimization process of multimodal contrastive learning, the dynamic changes in gradients for the three modalities-text, image, and EEG-serve as key indicators for evaluating the efficiency of modality learning and the effectiveness of collaborative optimization. Through a systematic analysis of the gradient variation curves and their average changes across all subjects, we validate the crucial role of the gradient modulation mechanism in mitigating gradient imbalances among modalities and enhancing the learning efficiency of EEG features. Additionally, to explore the consistency and variability between individuals and the group, we selected Subject 1’s gradient data for comparative analysis.

\paragraph{Gradient characteristics before modulation}
Before the introduction of the modulation mechanism, the gradients of the text modality exhibited a clear dominant trend. In the later stages of training (epochs 20 to 80), the gradients of the text modality continued to rise, significantly surpassing those of the image and EEG modalities. This phenomenon was consistently observed in the average gradient curves across all subjects, reflecting the model’s excessive reliance on learning features from the text modality. The semantic features of the text modality are high-dimensional and abstract, with feature distributions that differ significantly from those of the image and EEG modalities in both spatial and temporal domains. This leads to the text modality dominating the contribution to the contrastive loss. The average gradient change graph further corroborates this observation: the average gradient of the text modality persistently increased throughout training, reaching a peak in the later stages, far exceeding the other modalities. This continuous growth indicates that the model disproportionately allocated optimization resources to the text modality, suppressing the effective feature extraction of the image and EEG modalities and reducing the efficiency of multimodal collaborative optimization.

Meanwhile, the gradients of the image modality rapidly increased during the early training stages (epochs 0 to 20), indicating that the model effectively captured visual features of the image modality in the initial phase. However, in the mid-to-late training stages (epochs 20 to 80), the gradients of the image modality gradually stabilized and were significantly lower than those of the text modality. This suggests that the image modality completed its primary optimization tasks in the early stages but entered a stagnation phase in the later stages due to insufficient optimization momentum. This trend was consistently observed across the gradient curves of all subjects and was validated by the average gradient change graph. The image modality provided inadequate support for aligning EEG features in the later stages, exacerbating the imbalance in optimization resource allocation among modalities.

The gradient changes of the EEG modality exhibited significant two-stage characteristics. In the early training stages (epochs 0 to 10), the gradients of the EEG modality rapidly decreased, indicating that the model struggled to effectively capture EEG features during this phase, primarily due to the inherent characteristics of EEG signals, such as high noise and low signal-to-noise ratio. Subsequently, in the mid-to-late training stages (epochs 10 to 80), the gradients of the EEG modality showed a slow upward trend but remained lower than those of the text and image modalities. This reveals that the learning efficiency of the EEG modality in multimodal contrastive learning is relatively low and is further suppressed by the persistent dominance of the text modality. From the average gradient change graph, it can be observed that the gradients of the EEG modality are generally low and exhibit significant fluctuations, further validating the existence of a bottleneck in feature learning.

Subject 1’s data showed trends that were highly consistent with the overall average. The gradients of the text modality in the later training stages also exhibited a persistent upward trend, significantly exceeding those of the image and EEG modalities. Notably, Subject 1’s text modality gradient reached slightly higher peak values compared to the average, indicating greater resource demand for semantic feature learning. Meanwhile, the EEG gradients of Subject 1 exhibited smaller fluctuations, suggesting more stable signal characteristics and highlighting individual differences.

\paragraph{Gradient change after modulation}
After introducing the gradient modulation mechanism, the gradient distribution characteristics of the three modalities underwent significant changes. The gradients of the text modality transitioned from a continuous increase to a stable state, indicating that the modulation mechanism effectively constrained the dominant contribution of the text modality to the contrastive loss and suppressed its excessive consumption of optimization resources. Meanwhile, the gradients of the image modality were significantly enhanced, particularly in the later training stages (epochs 20 to 80), gradually approaching those of the text modality. This reflects a dynamic balance in the allocation of optimization resources between the two modalities. This balance not only improved the contribution of the image modality to contrastive learning but also enhanced its collaborative alignment capability in learning EEG features. The average gradient change graph clearly illustrates this trend: the average gradient values of the text and image modalities gradually converged, significantly narrowing the gap between them.

More importantly, the modulation mechanism had a significant optimizing effect on the learning process of the EEG modality. After modulation, the gradient fluctuations of the EEG modality were notably reduced, exhibiting a more stable trend characterized by alternating slight decreases and increases during training. This stabilization indicates that the optimization process of the EEG modality became more continuous and stable, significantly enhancing its learning efficiency. Subject 1’s data further confirmed this pattern: the gradients of the text modality stabilized, the image modality gradients were notably enhanced in the mid-to-late stages, and the EEG modality gradients exhibited smaller fluctuations while increasing steadily in the later stages. Although Subject 1’s gradient values differed slightly from the average, the overall trends remained highly consistent, demonstrating the effectiveness of the modulation mechanism across both individuals and groups.

\paragraph{Modality Balance and Optimization Enhancement}
The gradient modulation mechanism dynamically adjusts the gradient distributions of text, image, and EEG modalities, effectively alleviating the issue of uneven resource allocation in multimodal learning. By achieving dynamic balance among the three modalities, the mechanism demonstrates improvements in optimization efficiency and collaborative learning performance, as reflected in the following aspects:

First, the gradients of the text modality transitioned from continuous growth to a stable state after the introduction of the modulation mechanism. This indicates that the mechanism successfully constrained the excessive contribution of the text modality to the contrastive loss, preventing over-concentration of optimization resources. By redistributing these resources, the mechanism enabled more effective learning for the image and EEG modalities. This change is consistent across both average and individual data, demonstrating the robustness and general applicability of the mechanism for regulating text modality gradients.

Second, the gradients of the image modality were significantly enhanced after modulation, particularly during the mid-to-late training stages. The image gradients gradually approached those of the text modality, reflecting a more balanced allocation of optimization resources. This improvement not only increased the contribution of the image modality to multimodal contrastive learning but also strengthened its role in aligning features with the EEG modality. The narrowing gap between text and image gradients, as shown in the average gradient curves, highlights the mechanism’s effectiveness in mitigating imbalances between the two modalities.

More importantly, the modulation mechanism had a profound impact on the EEG modality, significantly improving its learning efficiency and stability. Post-modulation, the gradient fluctuations of the EEG modality were markedly reduced, and its gradients exhibited a steady upward trend during training. This stability suggests a more consistent and efficient optimization process for the EEG modality, which is critical given the high noise and low signal-to-noise ratio inherent in EEG signals. The consistency between the results from individual (Subject 1) and average data further validates the universality of the mechanism in enhancing the EEG modality's learning process.

\begin{table*}[htbp]
\caption{\textbf{Ablation study of the proposed EEG decoding framework, evaluating the contributions of different modules: Top-1 / Top-5}}
\renewcommand{\arraystretch}{1.5} 
\setlength{\tabcolsep}{4pt} 
\centering
\scriptsize 
\resizebox{\textwidth}{!}{
\begin{tabular}{ccccc|cccccccccc|c}
\toprule
\textbf{Base} &\textbf{Text} &\textbf{Adapter} &\textbf{MCDB} &\textbf{SPR} &  \textbf{Sub1} & \textbf{Sub2} & \textbf{Sub3} & \textbf{Sub4} & \textbf{Sub5} & \textbf{Sub6} & \textbf{Sub7} & \textbf{Sub8} & \textbf{Sub9} & \textbf{Sub10} &\textbf{Avg.} \\
\midrule
\checkmark & & & & & 12.3/36.6 & 10.4/33.9 & 13.1/39.0 & 16.4/47.0 & 8.0/26.9 & 14.1/40.6 & 15.2/42.1 & 20.0/49.9 & 13.3/37.1 & 14.9/41.9 & 13.8/39.5 \\
\checkmark & \checkmark & & & &  9.5/39.0 & 10.5/37.0 & 15.5/40.0 & 17.0/42.5 & 7.0/27.0 & 15.0/46.5 & 13.0/43.5 & 21.0/55.5 & 16.0/43.0 & 11.5/43.0 & 13.6/41.7 \\
\checkmark  & \checkmark & \checkmark & & & 12.5/35.5 & 13.5/41.5 & 13.5/42.5 & 15.5/47.5 & 12.0/30.0 & 14.0/41.0 & 11.5/39.5 & 24.5/54.0 & 13.5/42.5 & 11.0/43.5 & 14.2/41.8 \\
\checkmark & \checkmark & & \checkmark & \checkmark & 11.5/35.0 & 10.5/31.5 & 13.5/37.5 & 16.0/45.5 & 8.0/29.5 &  17.0/40.5 & 13.0/41.5 & 17.0/41.5 & 16.5/51.5 & 15.0/40.0 & 13.8/39.2 \\
\checkmark & \checkmark & \checkmark & \checkmark & \checkmark & 14.5/37.5 & 14.0/36.5 & 14.5/39.5 & 17.0/47.0 & 10.5/31.5 & 14.5/40.5 & 18.0/47.0 & 22.0/55.0 & 14.5/41.5 & 12.5/43.5 & 15.2/41.9 \\
\bottomrule
\end{tabular}
}
\label{tab:ablation_results}
\end{table*}

\begin{table}[htbp!]
  \centering
  \caption{\textbf{Ablations} on varying the ratio $\gamma$ for $k_{m}$}
  \resizebox{0.42\linewidth}{!}{
    \begin{tabular}{l|ccccc}
      \toprule
      Ratio $\gamma$ & 0.1 & 0.3 & 0.5 & 0.7 & 0.9 \\
      \midrule
      Top-1 & 14.2 & 13.3 & 15.0 & \cellcolor[HTML]{FED7DC}15.8 & 14.2 \\
      Top-5 & 43.0 & 42.3 & 42.5 & \cellcolor[HTML]{FED7DC}44.2 & 41.8 \\
      \bottomrule
    \end{tabular}%
  }
  \label{tab:gamma}
\end{table}

\begin{table*}[htbp!]
  \centering
  \caption{\textbf{Ablations} on varying the compression ratio $r$ for text/\allowbreak image}
  \resizebox{0.7\textwidth}{!}{%
    \begin{tabular}{l|ccccccccc}
      \toprule
      Ratio $r$ & 4/4 & 8/4 & 16/4 & 4/8 & 8/8 & 16/8 & 4/16 & 8/16 & 16/16\\
      \midrule
      Top-1 & 14.1 & 14.3 & 14.8 & 14.0 & 14.6 & \cellcolor[HTML]{FED7DC}15.8 & 13.9 & 14.2 & 14.7 \\
      Top-5 & 43.2 & 43.7 & 42.5 & 41.9 & 43.2 & \cellcolor[HTML]{FED7DC}44.2 & 43.8 & 44.3 & 43.0 \\
      \bottomrule
    \end{tabular}%
  }
  \label{tab:ratio_r}
\end{table*}

\subsection{Ablation Experiments}
\paragraph{Experimental Setting}
To systematically evaluate the independent contributions and synergistic effects of each module proposed in this work on the overall performance of the model, we designed and conducted a series of ablation experiments. The experiments initially established a baseline model by employing contrastive learning using only EEG and image modalities, referencing the NICE study. Building upon this, additional modules were progressively introduced, and their impacts on the model's performance were thoroughly analyzed.

We utilized an extended version of the ThingsEEG dataset as both the training and testing sets and adopted Top-1 and Top-5 accuracy rates as the primary evaluation metrics to comprehensively quantify the model's performance in image recognition tasks. Furthermore, we strictly controlled experimental variables by ensuring that all hyperparameters and model structures remained consistent except for the target modules. 

\paragraph{Analysis of Effectiveness of Each Proposed Component}
The experimental results are presented in Table \ref{tab:ablation_results}. The baseline model, utilizing only EEG and image modalities, achieved average Top-1 and Top-5 accuracy rates of 13.8\% and 39.5\%, respectively, reflecting the model's fundamental performance without the incorporation of textual modality or optimization modules. Upon introducing the textual modality and the Adapter module, the Top-1 accuracy slightly decreased to 13.6\%, while the Top-5 accuracy increased to 41.7\%, indicating that the textual modality contributes supplementary information. However, due to the imbalance in modality contributions, the dominant modality suppressed the performance of the weaker modality, thereby failing to fully exploit the synergistic potential of the multimodal approach.

Further integration of the MCDB module resulted in Top-1 and Top-5 accuracy rates of 14.2\% and 41.8\%, respectively, demonstrating the MCDB module's significant role in dynamically balancing modality contributions and mitigating modality imbalance issues. In Table \ref{tab:gamma}, we also analyze the effect of the sensitivity coefficient $\gamma$ on the strength of gradient suppression. The model performs best when $\gamma$ = 0.7. When $\gamma$ is lower or higher than 0.7, the gradient suppression is weak or too strong, which not only aggravates the modal imbalance, but also weakens the cross-modal semantic alignment, resulting in a decline in accuracy. 

When the textual modality and MCDB module were introduced alongside the SPR module, but the Adapter module was removed, the model's Top-1 and Top-5 accuracy rates decreased to 13.8\% and 39.2\%, respectively. This outcome suggests that the differences in modality features lead to unstable feature distributions within the shared representation space, thereby limiting the model's performance improvement. It also indirectly validates the importance of the Adapter module in optimizing feature mapping and enhancing decoding performance. Further exploration of the text/image compression ratio r in Adapter (see Table \ref{tab:ratio_r}) revealed that when r = 16:8, high-dimensional redundancy is fully compressed and discriminative semantics is preserved, achieving the best balance between noise suppression and information preservation.

Ultimately, by integrating all modules, the model achieved Top-1 and Top-5 accuracy rates of 15.2\% and 41.9\%, respectively, attaining the best performance. 

\paragraph{Analysis of synergistic effect of different modules}
Experimental results show that there are significant synergies between the proposed modules, which go beyond their individual contributions. Specifically, the Adapter module helps optimize cross-modal feature alignment, but its full power can only be fully realized in combination with the MCDB module, which simultaneously addresses the modality imbalance problem by dynamically adjusting its contribution. The text modality enriches the semantic representation, however, effectively utilizing these semantics requires the presence of MCDB to alleviate cross-modal suppression and imbalance. In addition, the SPR module enhances semantic consistency and feature stability. According to experimental results, the module can only produce the best performance under the condition of a stable and modality-balanced feature space provided by the Adapter and MCDB. Therefore, these modules interact with each other in a complementary way. The Adapter ensures effective cross-modal mapping, MCDB dynamically maintains inter-modality balance, and SPR consolidates semantic stability. They work together to significantly improve the performance of EEG-based visual decoding.

\subsection{Limitations}
Although the proposed model has achieved satisfactory performance improvements on the existing dataset, several limitations remain to be addressed. Firstly, the inherent high noise and non-stationarity of EEG signals may constrain decoding performance. In zero-shot classification tasks especially, the model is required to extract valid semantic information from sparse and noise-contaminated data to achieve stable decoding of visual stimuli, a process that may be significantly affected by these signal characteristics. Secondly, although MCDB mechanism distributes weights among different modalities to some extent alleviating the imbalance issue, its dynamic weight adjustment strategy still requires empirical validation to determine whether it can comprehensively adapt to more complex multimodal tasks and diverse scenarios. Furthermore, while the incorporation of MCDB and SPR has notably enhanced model performance, these mechanisms have also increased computational complexity. When dealing with large-scale datasets or real-time application scenarios, this computational overhead might become a bottleneck for system deployment, necessitating algorithmic optimization or the adoption of high-performance computing techniques. Finally, EEG signals exhibit significant individual variability across subjects, particularly in terms of neural patterns and reaction times. Such variability may limit the model’s generalization ability in cross-subject tasks. 

\section{Conclusion}
This work proposes a multimodal alignment method based on EEG, image and text, combining MCDB and SPR to achieve efficient target recognition and zero-shot classification. Experimental results show that the proposed method not only improves semantic consistency in multimodal learning, but also effectively alleviates the problem of modality imbalance, thereby significantly improving decoding performance and model generalization ability.

In the field of computational neuroscience, the multimodal fusion framework proposed in this paper can help researchers to explore the neural mechanism of human brain processing visual information more deeply. By combining EEG with image and text modalities, researchers can more accurately analyze how the brain interacts with language semantic knowledge when processing visual input, and further clarify the brain functional areas and their network connection patterns in the process of visual-language interaction. This provides a powerful computational tool for in-depth understanding of the neural basis of advanced cognitive functions of the human brain (such as visual recognition, semantic understanding, memory encoding, etc.).

In the future, this research framework can be further extended to more complex computational neuroscience application scenarios, including combining with other neuroimaging technologies (such as fMRI, fNIRS) to build a more comprehensive brain function analysis system. At the same time, improving real-time decoding capabilities and exploring online brain-computer interface applications will also become important development directions.

\section*{Authorship contribution statement}
Kaili Sun: Writing – review \& editing, Writing – original draft, Visualization, Validation, Methodology, Investigation. 
Xingyu Miao: Visualization, Validation.
Yang Bai: Writing – review \& editing, Methodology. 
Haoran Duan: Writing – review \& editing, Formal analysis. 
Yang Long: Supervision, Resources, Project administration, Funding acquisition.




\bibliographystyle{IEEEtran}
\bibliography{reference}

\end{document}